\documentclass[times,twocolumn,final]{elsarticle}

\usepackage{displa}
\usepackage{framed,multirow}

\usepackage{amssymb}
\usepackage{latexsym}

\usepackage{url}
\usepackage{xcolor}
\usepackage{amsmath,amssymb,amsfonts}
\usepackage{algorithmic}
\usepackage{framed,multirow}
\definecolor{newcolor}{rgb}{.8,.349,.1}

\usepackage{hyperref}

\journal{Displays}

\begin{document}

\verso{Pathirage N. Deelaka }

\begin{frontmatter}

\title{{Neural Artistic Style Transfer with Conditional Adversarial Networks\tnoteref{tnote1}}}

\author{\snm{Pathirage} N. Deelaka\fnref{fn1}}
\address{Department of Computer Science and Engineering, University of Moratuwa, Moratuwa 10400, Sri Lanka}


\received{1 May 2013}
\finalform{10 May 2013}
\accepted{13 May 2013}
\availableonline{15 May 2013}
\communicated{S. Sarkar}

\begin{abstract}
A neural artistic style transformation (NST) model can modify the appearance of a simple image by adding the style of a famous image. Even though the transformed images do not look precisely like artworks by the same artist of the respective style images, the generated images are appealing. Generally, a trained NST model specialises in a style, and a single image represents that style. However, generating an image under a new style is a tedious process, which includes full model training. In this paper, we present two methods that step toward the style image independent neural style transfer model. In other words, the trained model could generate semantically accurate generated image under any content, style image input pair. Our novel contribution is a unidirectional-GAN model that ensures the Cyclic consistency by the model architecture.Furthermore, this leads to much smaller model size and an efficient training and validation phase.
\end{abstract}


\end{frontmatter}


\section{Introduction}
Imposing a style on an image is one of the most laborious tasks in graphic designing. Most of the time, this process is handled by a skillful graphic designer, and it will take hours to finish one image with good quality. Using a neural style transfer (NST) model like \cite{b1} is not popular among the computer graphics community due to several reasons. Since a model is specialized for the trained style, a simple application that supports several style transfers would be significant in terms of storage, considering the size of one NST model. Furthermore, the NST model imposing alias artifacts on the input image makes the model less reliable. Our first goal in this paper is to develop a NST model that supports more than one style to transfer. The second goal is to introduce a reliable NST model that imposes only general features related to a style.

\begin{figure*}[!t]
  \centering
  \includegraphics[width=0.98\textwidth]{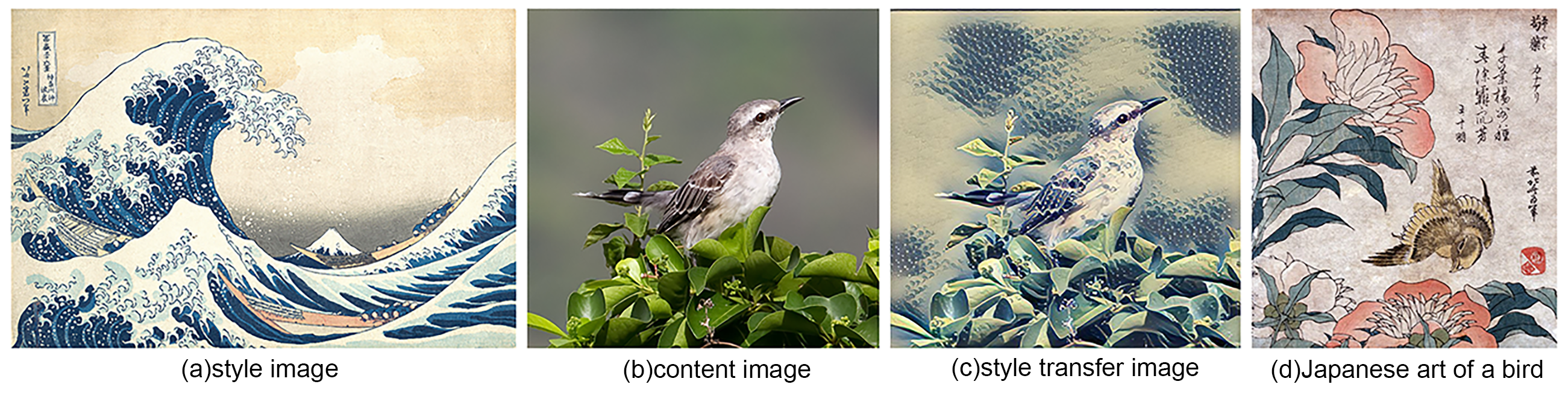}
  \caption{ The generated image (c), which is generated for the content image (b) using the CNN based NST model trained on the style image (a), shows to be different from the image (d), which represents a Japanese artwork with the same semantics of the content image (b).}
  \label{fig:cover}
\end{figure*}

Motivated by the Generative Adversarial Network (GAN) \cite{b4, b5, b14} based image-to-image translation and style transfer literature \cite{b7}, we upgrade the convolutional neural network (CNN) based generator architecture into a conditional generative adversarial network(cGAN) based image synthesis process. Both proposed models have two independent discriminators to generate content based adversarial loss and style based adversarial loss. The concept introduced in the \cite{b7} on understanding and controlling latent space to impose desirable features on generated images lay the fundamental idea to the second model we introduce in this paper. Using a mapping network introduced in \cite{b7} for matrix learning \cite{b23} make the encoder learn style features like in Natural Language Processing semantic classification model generates feature embedding for tokenized words.

CNN based NST models introduced so far \cite{b1, b3, b7} are models that train thousands of content images only against a single image which will define the transform style of the model trained. The objective function of this supervised training was the weighted sum of content loss and style loss. Style loss function calculates the norm of correlation matrix between style and input images. Images generated using the above conventional NST embeds alias artifacts from the style image and changes the color palette of the generated image significantly from the content image, as shown in Fig.~\ref{fig:cover}. Since neural artistic style translation is an unpaired image-to-image translation, by definition, it makes NST unfitting for the Pix-to-Pix GAN \cite{b3} model. Even though StyleGAN \cite{b7} model is designed on an unpaired image-to-image translation between two pixel spaces, it uses the same set of features to generate the image from a separate pixel space.

In this paper, we explore GANs in the NST, where just as the StyleGAN model generates an image with the mixture of both input image features. The proposed model extracts style features from the style image and object features from the content image and generates a new image with objects from the content image painted in the style of the style image. For such a case, we propose a GAN model that can be used for NST, which can support unpaired image-to-image translation on independent input pixel spaces. Furthermore, the proposed model extracts and generates a different latent space on both style and content image, and the pixel space of the generated image differs from both input images’ pixel spaces. 

We introduce a new GAN architecture with two discriminative headers in the first approach. In the second approach, we introduce an advancement to the architecture of the first approach, which makes the training separate parameter spaces for discriminators disappear. In the revealing network architecture, the encoder sub-models in the generator work as the discriminative headers in the training phase. In this model training process, generator training and discriminator training optimize the same parameter space.

Conditional adversarial network architecture has been extensively upgraded over the past few years to support various image synthesis tasks in the computer vision field. Even though there were improvements in CNN based NST models such as \cite{b3, b7}, they were based on CNN supervised learning as in \cite{b1}. The major contribution of this paper is to introduce a new cGAN architecture where the discriminator and encoder sub model of the generator share the same parameter space in the training time.  The generator model has been made more reliable and consistent in style transferring as in the CycleGAN paper \cite{b6} introduced.

\begin{table*}[!h]
 \centering
\begin{tabular}{ |p{5cm}|p{1cm}|p{1cm}|p{1cm}|p{1cm}|p{1cm}|p{1.5cm}|p{1.5cm}| } 
\hline
 & Svoboda \cite{b28} & Gao \cite{b29} & Chen \cite{b31} & Gatys \cite{b1} & Zhang \cite{b30} & Appr. 1 & Appr. 2 \\
\hline
Semi-supervised learning & $\surd$ & $\surd$ & \ & \ & $\surd$ & $\surd$ & $\surd$ \\
\hline
does not require paired samples to train & \ & $\surd$ & \ & $\surd$ & \ & \ & $\surd$\\
\hline
trained model support more than one style & \ & \ & $\surd$ & \ & $\surd$ & $\surd$ & $\surd$ \\
\hline
preserve original image color & \ & \ & $\surd$ & $\surd$ & \ & \ & $\surd$ \\
\hline
does not introduce alias artifact in generated image & $\surd$ & $\surd$ & \ & \ & $\surd$ & $\surd$ & $\surd$ \\
\hline
\end{tabular}
\caption{Table of comparison}
\label{table:1}
\end{table*}

\section{Related Studies}
A neural style transfer paper \cite{b1} by Leon et al. first introduced the concept of neural style transformation under image synthesis in the deep learning based computer vision field in 2015. Since then, it has been prevalent in research because of its mainstream appeal. In \cite{b7}, the authors have introduced an extension to the \cite{b1} to make the image synthesis process more efficient and faster in the NST process. Also, the \cite{b3} color histogram matching algorithm has been used on top of \cite{b1} model architecture to preserve the original light condition in the input image.

In \cite{b28, b29, b30} they have used semi-supervised learning (GAN) to achieve more controllability in the style transferring process, produce clear and detailed images, and disentangled feature \cite{b28} from both content and style feature spaces. However, as shown in the \ref{table:1}, most research studies in NST have been conducted under the concept of a model per style. 

Use of GAN \cite{b4, b34} in image synthesis, image super-resolution, image editing, and representation learning is very popular. Google pix-to-pix translation \cite{b5} paper has introduced a U-net based image to image translation model that was trained with GAN. It showed impressive performance in semantic segmentation, semantic labeling, map translation [aerial photo into graphics], edge and boundary detection, and in image-to-image translation tasks such as thermal to color translation. However, the pix-to-pix model could only have a single input. This restricted the use of the pix-to-pix model for the style transfer model. Additionally, the model requires to be fed with original and translated images as a pair. 
However, it is not practical to generate transferred images beforehand like in other image synthesis areas in style transfer.  For example, we can manually annotate and pair images in semantic segmentation. Jun-Yan Zhu et al. \cite{b6}, has introduced a consistent and well defined method for bijective validation of image translation between two pixel spaces. In \cite{b6}, they have introduced image translation between two domains with unpaired initial distributions.

In design, the CycleGAN model takes a single image input, making CycleGAN unsuitable for direct use in style transferring. In 2019, StyleGAN model \cite{b7} paper has introduced an effective model to combine features from two different input images and generate a new image with features of input images. Since that paper targeted synthesizing human faces,  both images fed into the model were human images. Therefore, they have proposed only a single encoder model (Mapping Network) \cite{b7} for feature extraction. Furthermore, since features in human faces were explicitly localized and acceptable network architecture for global features like texture extraction differ from the Mapping Network architecture.

The Wavelet Convolutional neural network model \cite{b8} has been proposed to classify images based on texture features in wavelet representation. The input image was first operated on a multi-resolution image decomposition method called spectral analysis. However, since the wavelet transformation maps pixels into a spectral space, it cannot be used for the generator's style/texture encoding model In \cite{b10, b11}.  They have empirically proven that convolutional models with residual connections can be used for texture classification tasks. In \cite{b10} they have claimed that pre-trained CNN models on ImageNet \cite{b23} dataset have the potential to recognize texture features as well as the shapes such as clothes in complex images.

As a part of the style transfer process, we have to extract style/texture from the style image. The style feature of an artwork is represented by different factors such as the medium used to paint, the surface, and the artist's personality. The paper published by Yang Lia et al. \cite{b37} introduced the GAN-based method to control the style ingrained on the aspect of the continuous flow of color gradient to incorporate with brush stroke patterns in artworks.

\section{System Model and Implementation}
In the first approach, we propose a GAN model where the discriminator has two parallel headers to independently generate adversarial results on content and style. In the same approach, we synthesize the training image dataset by using the NST \cite{b7} model to generate paired style transfer images. We advance the cGAN model architecture and the training process for the model to understand the class related to the style image and extract core features from the content model. Then the model imposes general features of the identified style class to the generated image. Specifically, using the second method, we can generate images that are not affected by alias feature presence like in Fig.~\ref{fig:cover}. 

Since the style encoding submodule is optimized under the metrics learning objective function for image similarity search, given that the model is trained under a higher variation of style classes \cite{b27}, the trained model will support any art style irrespective of the style presence in the training time. For the approach one training, we synthesize a dataset in the form of an image matrix. In the matrix, every column denotes a content image from the Multi-Salient-Object (MSO) dataset \cite{b26}. All other columns contain style transfer images of content images from MSO related to the row and style related to the particular column.

\subsection{Approach \#1 : GAN with dual headed discriminator}
One of the Foundation work for this approach has been done by the pix-to-pix paper \cite{b5}, as a cross domain image transformation model with great qualitative and qualitative accuracy in the generated images. However, the architecture's nature is not directly compatible with the NST for two reasons. The first reason is that the GAN architecture must support two images as the inputs, and the second reason is that the NST is an unpaired image to image transformation in contrast to the pix-to-pix model. 

\subsubsection{Objective of approach \#1}
the cGANs \cite{b23} generator learns a mapping from the input image domain $x$ \& noise vector $z$ to paired image domain $y$, $G:\{x,z\} \xrightarrow{} y$, to produce "real" images. While  the discriminator $D$ learns to distinguish "real" images from "fake" images generated by the generator $G$. 

The objective of cGAN can be expressed as.
\begin{multline}
    \mathcal{L}_{cGAN}(\mathcal{G},\mathcal{D}) = \\
    \mathbb{E}_{x,y}[\log(\mathcal{D}(x,y))]\ +\ \mathbb{E}_{x,y}[\log(1 - \mathcal{D}(x,\mathcal{G}(x,z)))]
\end{multline}
Here, the generator $G$ tries to minimize this objective against the adversarial discriminator  $D$ tries to maximize the adversarial loss.\\
i.e. 
\begin{math}\mathcal{G}^* = \arg \min_\mathcal{G} \max_\mathcal{D} \mathcal{L}_{cGAN}(\mathcal{G},\mathcal{D})
\end{math}
\\
The proposing model has two distinct discriminators against one generator, and here onward we use rGAN as the short form for this proposing model. Each discriminator trains separately. Weighted sum of adversarial loss defines as the objective of rGAN can express as;
\begin{multline} \label{loss_1}
    \mathcal{L}_{rGAN}(\mathcal{G},\mathcal{D}_s,\mathcal{D}_c) = \\ \alpha.  \mathcal{L}_{cGAN}(\mathcal{G},\mathcal{D}_s)\ +\ (1-\alpha). \mathcal{L}_{cGAN}(\mathcal{G},\mathcal{D}_c)
\end{multline}

As stated in the \cite{b5} we added $L$1 loss for perceptual clarity of the generating image. Thus, the final objective is,
\begin{equation} \label{obj_1}
\mathcal{G}^* = \arg \min_\mathcal{G} \max_{\mathcal{D}_s} \max_{\mathcal{D}_c} \mathcal{L}_{rGAN}(\mathcal{G},\mathcal{D}_s,\mathcal{D}_c)\ +\ \lambda.\mathcal{L}_{l1}(\mathcal{G})
\end{equation}

A pairwise marginal ranking loss used for the objective function in the content to assess synthesis image quality \cite{b32} of generated images.

\subsubsection{Network Architecture of approach \#1}
\begin{figure}[!t]
 \centering
  \includegraphics[width=\linewidth]{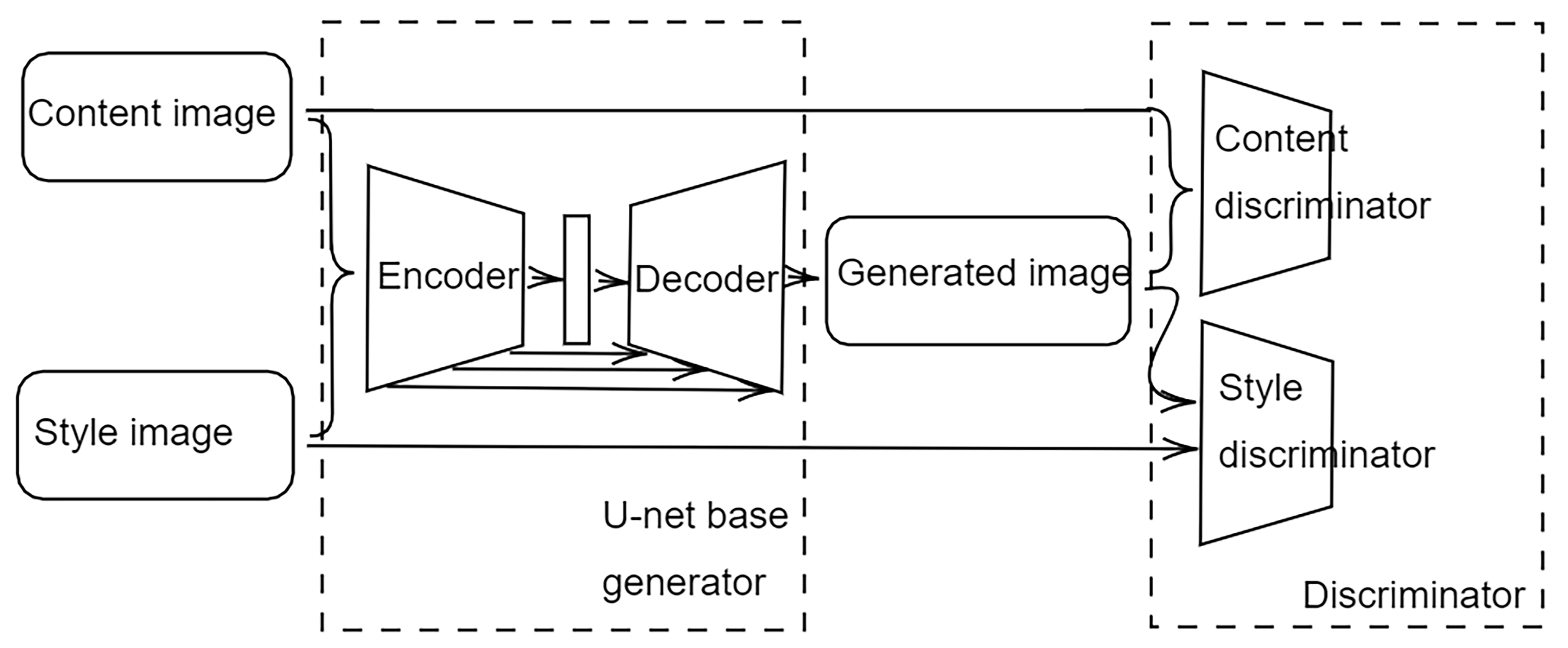}
  \caption{The GAN architecture of approach \#1.}
  \label{fig:app_1_gan}
\end{figure}

The skeleton of the GAN model is adopted from the paper \cite{b5} with an added style discriminator header and channels in the encoder to support the style image. Generator and Content discriminator use standard convolutional modules [Conv $\xrightarrow{}$ Batchnorm$\xrightarrow{}$ ReLU]. However, the style discriminator is designed by the wavelet convolution neural network. The complete training procedure diagrammed in Fig.~\ref{fig:app_1_gan}

\paragraph{\textbf{Generator with skip connection} -}
The generator model aims to extract object features from the content image and local-global fused features from the style image. Subsequently, those features are encoded from the input images' pixel space into bottleneck layers' latent vector space. The decoder model reconstructs the style transferred image up to the same size as the input. The bottleneck vector size is set to 64 as defined in \cite{b12}.

Apart from the two inputs taken by the encoder model, the generator architecture is a U-Net architecture \cite{b13}. Since the U-Net allows information sharing from low level features of the encoder at the image reconstruction step, it gives a chance to fuse low level features related to the texture of the style image. Furthermore, As in the \cite{b5} states, U-net residual architecture supports reducing blurriness in the generated image.

\paragraph{\textbf{Markovian discriminator [content discriminator]} -}
The exact discriminator designed in \cite{b5} is used due to several reasons. Since generated images are mixed with style image features, it is crucial to ensure that local image patches are real. PatchGAN \cite{b5} penalizing structure at the scale of patches ensures the content image is not embedded with alias artifacts, further allowing the generation of a more semantically accurate image. The discriminator considers the image a Markov random field with plate per patch. Therefore, it indirectly supports having original colors in the generated image because each patch is considered independent from surrounding patches.

\paragraph{\textbf{wavelet CNN discriminator [style discriminator]} -}
Theoretically, the style of an image cannot be successfully captured by a sequential CNN model because CNN targets capturing local features and transforming those features until it reaches a complexity that the fully connected layer will support. As stated in \cite{b10, b36}, even a CNN model with a feature fusion like residual CNN \cite{b19}, the inception model \cite{b24} can be used for texture extraction tasks. The wavelet convolution neural networks begin with the wavelet transformation layer, which decomposes pixel space into spectral space in several resolutions using frequency statistics.  Then in the Convolutional Neural Network potion, we down-sample different resolution results from the wavelet layer with channel-wise adding. In essence, this model acts as a global-local feature fusion to extract style from the images. 

The paper \cite{b8} introduced a state-of-the-art texture extraction architecture based on the wavelet transformation. The style discriminator header is implemented using wavelet CNN in this approach. Using this model in the generator as a style encoder header adds unexpected hues into the generated image and partitions the generated image by light black stripes. We stipulate that features cause these behaviors in the CNN followed by a wavelet transformation layer in spectral space.

\subsubsection{Optimization and inference}

As suggested in the original GAN \cite{b4} we alternate training between generator model and discriminator model to get the generator to converge into a point where loss is minimized, and discriminators damps on equilibrium level. Instead of feed forwarding real and fake sample sets separately over the discriminator's heads, with the motive of achieving a more robust gradient descent, we concatenate real and fake sample sets and feed them into the discriminator after a shuffling. More robust gradient descent can be achieved using this approach because, in the previous method, parameters of Batch Normalization layers radically changed due to real and fake sample twist feeding, but in this approach, batch normalization layers will be in a more stabilized position.

We used a non-reference loss function - L1 norm as the image reconstruction loss in the generative model, mainly because the L1 loss preserves color \& luminance and equally weights the loss regardless of local structure. Furthermore, L1-loss does not penalize pixel value difference thoroughly. Therefore, it gives a chance to add style image color variations and features on the content image. We have also experimented on a reference loss function called multi-scale structural similarity index (SSIM) \cite{b15}, which is sensitive to local structure. Also experimented using the MixLoss function \cite{b16} which is a weighted combination of SSIM and L1 losses as the reconstruction loss. However, the generated images in both experiments gave unexpected hues and noise.

In the content discriminator header, we have used binary cross-entropy loss as suggested in \cite{b5}. We used an instance normalization layer instead of batch normalization for a more robust and smooth training curve in the style discriminator header. For the style loss function, we have implemented a slightly modified version of the pairwise marginal loss function \cite{b17} to support both positive and negative pairs at the same time.

Since the objective of the GAN base of paired image transformation, we have used it as stated in Method for model training. From that image matrix, we synthesize the training dataset in the form of triplets; [content image, style image, style transfer image]. We fed the batched dataset into the model training pipeline.

\subsection{Approach \#2 : GAN with discriminatory encoder sub models}

\begin{figure}[!t]
 \centering
  \includegraphics[width=\linewidth]{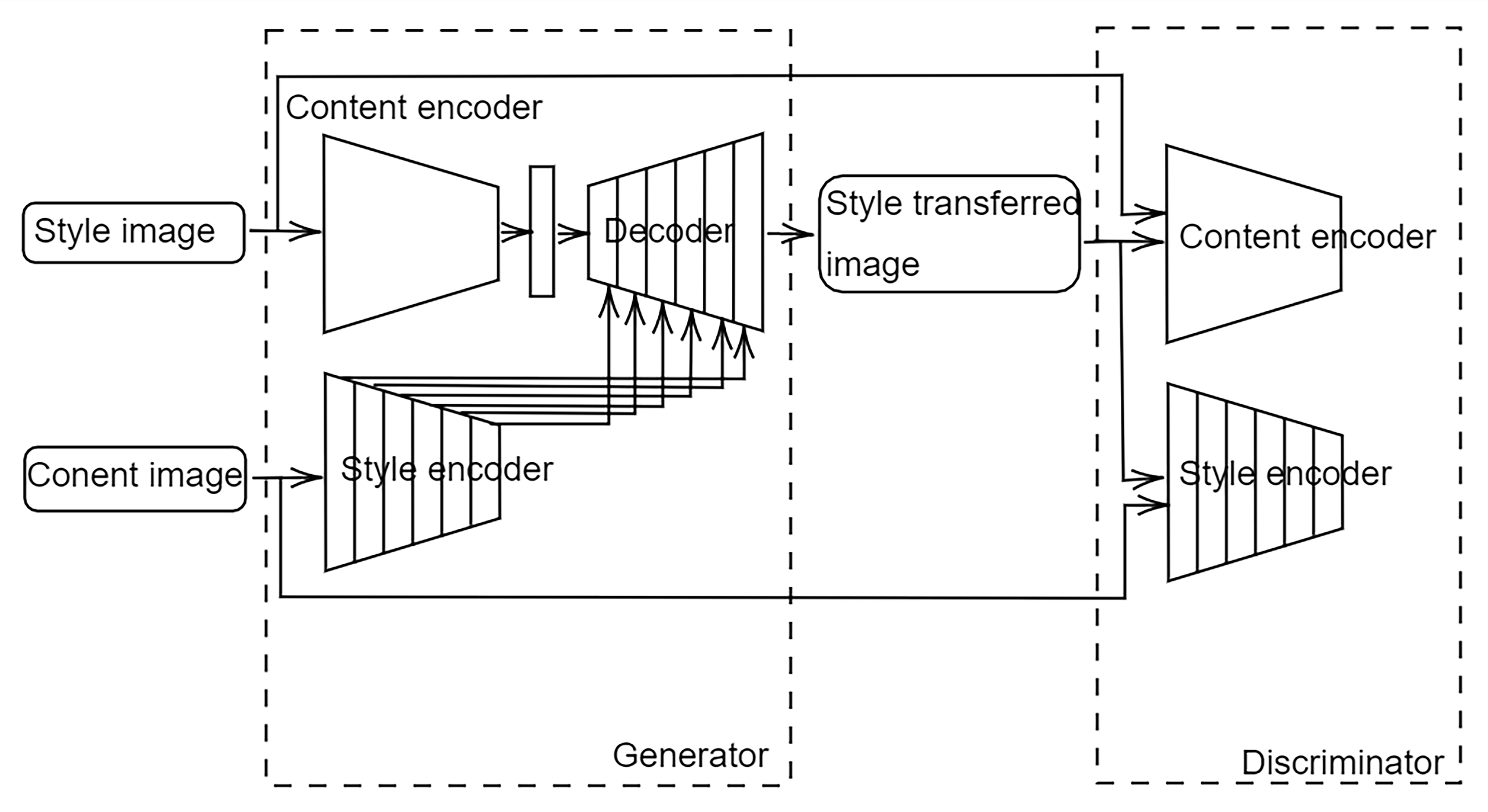}
  \caption{The GAN architecture of approach \#2.}
  \label{fig:app_2_gan}
\end{figure}

Approach \#1 targeted designing a model that can transfer images into more than one style. Therefore, to support a new style, we had to train a new CNN-based NST model such as the fast style transfer \cite{b7} model on a chosen set of styles. The style transfer images from the trained model on content images will be composed into an image matrix in the approach \#1. After that, the GAN in approach \#1 will be trained on the composed image matrix with randomized image augmentations on the pre-processing pipeline. Therefore, even if the model generates semantically accurate style transfer images on several styles, the complicated process of training the model for a new style is a downside of the approach \#1.

Also, in the paper \cite{b1} the original image color palette remains consistent because the model proposed by Approach \#1 still learns style from a single image, and it restricts the color palette of the generated image. As we discussed, making a style represented by a single image, in reality, is not accurate and could introduce many unnecessary features into the generated image. For example, in Fig.~\ref{fig:cover}, (third image) generated image embeds many alias artifacts such as the presence of waves instead of blurred background in the reference image (second image) and curly sprinkles like features in bird feathers. Furthermore, the fourth image in Fig.~\ref{fig:cover} shows a scene of a bird with leaves, and that image significantly differs from the generated image (third image). The generated image has embedded a blueish color palette like in the style image (first image of Fig.~\ref{fig:cover}) of the model. Both \cite{b1} and the proposed GAN model generates images that show the style image feature in the generated image. 

Besides the ample parameter space in the generative model, a discriminator model with a large parameter space degrades the gradient flow in the training phase. As a result of this ample parameter space [order of $10^5$], training the model to get at least moderate results requires a relatively large data set [order of $10^4$]. Also, it is hard to expect to converge the generative model into an optimal point while the discriminator oscillates over the proper equilibrium level.
Because adversarial loss employs to assess the generated samples against “real” samples, it is hard to guarantee that the generator learns for optimal transformation operation without collapsing. Further defining evaluation metrics on image reconstruction for style transfer is even more challenging. Also, evaluating model training by a metrics function does not always guarantee model learning. All the facts above indicate that using conventional ways to build a model that understands style transfer on unpaired data is not an accurate and reliable method.

In this approach, we are going to introduce a new model training process as the solution. The starting point is to drop the concept of training separate discriminator models. In this approach, the discriminator is not fed with fake samples as in GAN training. In this approach, we define two separate encoder models for content feature extraction and style feature extraction. As shown in Fig.~\ref{fig:app_2_gan}, we fed extracted features into the decoder to generate style transfer images. After that, we use two encoder models in the discriminator to generate an adversarial loss.

\subsubsection{Objective of approach \#2}
Even though we have adopted cGAN in this approach, the training phase will not be the same as in cGAN training. In this approach both generator and discriminator trains to minimize the objective, basically because the discriminator is a part of the generator. The objective function of $rGAN$ can express as;

\begin{multline} \label{loss_2}
\mathcal{L}_{rGAN}(\mathcal{G},\mathcal{D}_s,\mathcal{D}_c) = \\
\alpha. \mathcal{L}_{cGAN}(\mathcal{G},\mathcal{D}_s)\ +\ (1-\alpha). \mathcal{L}_{cGAN}(\mathcal{G},\mathcal{D}_c)
\end{multline}

same as in the Approach \#1, but the final Objective \ref{loss_2} is different from \ref{loss_1}.

\begin{equation} \label{obj_2}
\mathcal{G}^* = \arg \min_\mathcal{G} \min_{\mathcal{D}_s} \min_{\mathcal{D}_c} \mathcal{L}_{rGAN}(\mathcal{G},\mathcal{D}_s,\mathcal{D}_c)\ +\ \lambda.\mathcal{L}_{l1}(\mathcal{G})
\end{equation}

As in \cite{b6}, the objective is defined to validate that the transformation is learned by the rGAN. Since here we are using the same encoder models in the discriminator, it will validate the whole training process. Because, in the first step, encoder models train as a discriminator model to minimize loss of feature extraction. Then, in the second step, use those encoder models  in the generator model to adversarial train the decoder with encoders to minimize adversarial  loss. Those two steps repeat over the data set as same as in the cGAN training.

Especially, in the discriminator training process we do not feed "fake" samples as in cGAN training. This is because encoders are supposed to train in a way they generate optimal latent vectors instead of classification. In this architecture both discriminator models output encoding vectors as they are used in the generator. We used the pairwise marginal loss function for the objective of the content encoder training in the discriminator. Style encoder trained under metrics learning objective for image similarity search. 
\begin{equation} \label{ds_1}
    H_{i,k}\ =\ \frac{1}{\gamma}.\sum_{j} xa_{i,j}.xs_{j,k}
\end{equation}

\begin{equation} \label{ds_2}
    \mathcal{L}(xa, xs, ys)\ =\ -\sum_{i} ys_i.\log(H_i)
\end{equation}

In above equations \ref{ds_1} \& \ref{ds_2}, $xa$, $xs$ and $ys$ denote anchor sample batch, style sample batch and style label vector respectively. $\gamma$ used for regularizing constant in Gram metrics. Use the Gram metrics in \ref{ds_1} for sparse categorical cross-entropy in \ref{ds_2}.

In essence, encoder models are trained to generate optimal features while the generator model is trained to generate images that minimize weighted sum of content, style and l1 loss as shown in the equation \ref{obj_2} by using trained encoders.

\subsubsection{Network architecture of approach \#2}
The entire network consists of mainly three parameter spaces called a content encoder, style encoder, and decoder, as shown in Fig.~\ref{fig:app_2_gan}. Content encoder and style encoder outputs latent vectors of content image and style image, respectively. The decoder uses those encoded features in the generative model to generate style transfer images. The discriminative model uses the content encoder and style encoder to generate adversarial losses to train.

\paragraph{\textbf{Content encoder} -}
Content encoders take an image and output an encoded latent vector [of length 32]. An encoder is created by a set of sequential modules in the form of [Conv $\xrightarrow{}$ Batchnorm$\xrightarrow{}$ ReLU]. The content encoder is trained on pairwise marginal loss in the training process. As an adoption model of StyleGAN \cite{b7} synthesis network here, we do not directly feed encoded features over the bottleneck to the decoder model. Instead, we feed features using skip connections from the content encoding models intermediate CNN modules to intermediate up-sampling CNN modules in the decoding model. 

In the training process, the content encoder is trained in the way a face-recognition model is being trained \cite{b12}. Instead of using triplet loss \cite{b17}, here we are using the sum of pairwise marginal loss on the positive sample and negative sample in each training step. The content discriminator is fed by random sample pairs in the training process, where positive and negative pairs 	are randomly generated in a way each has a 0.5 probability. Also, instead of using semi-hard ranking loss functions like triplet loss, hard ranking loss functions, namely the pairwise marginal loss function in the discriminator make the gradient descent more robust. As stated above, we can guarantee that the content encoder model is trained to do expected tasks because the same encoder model is trained on content encoding vector classification by using a ranking loss function in the discriminator training step. Furthermore, we have assessed the accuracy of the content discriminator network and objective function on a separate data set.
\begin{figure}[!t]
 \centering
  \includegraphics[width=\linewidth]{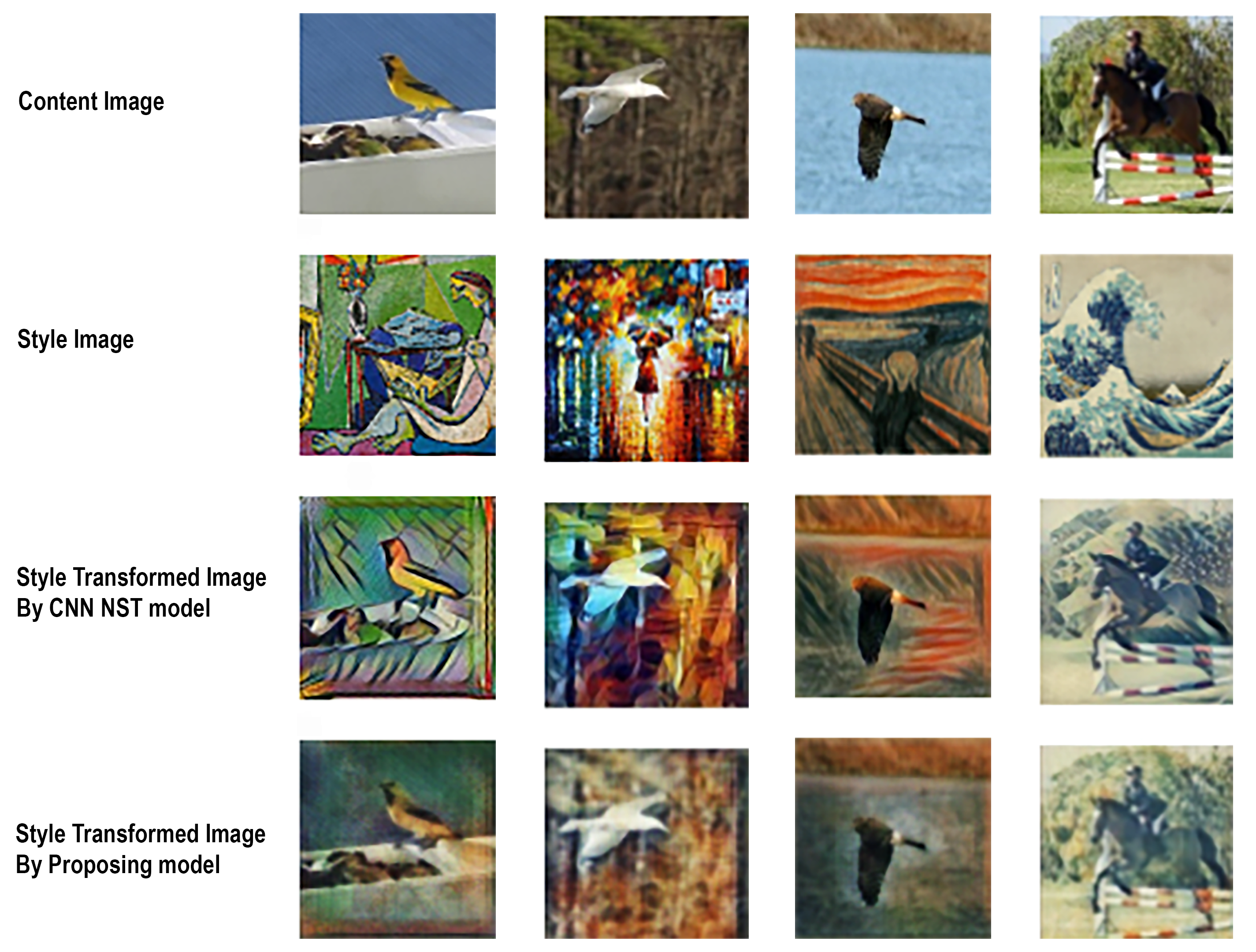}
  \caption{Approach \#1 GAN model evaluation image matrix.}
  \label{fig:app_1_eval}
\end{figure}

\paragraph{\textbf{Style encoder} -}
The major concern here is that encoder vectors must show locational proximity to indicate semantic similarity in the original style when transformed into UMAP, PCA base encoded vector spaces. For example, the pencil drawing's encoder vector position in point cloud must have proximity to a pencil sketch image rather than an oil painting. In essence, the style encoder is trained to yield approximately similar encoding vectors for similar style images. The task appears to be like embedding a vector training process but defining an objective function to train and using that in an adversarial loss function causes many issues. Also, the use of classification objective function to train is not accurate in theory and empirically. As stated in CycleGAN paper \cite{b6} “building $y$, $G: x\xrightarrow{} y$, mapping does not guarantee that an individual input x and an output y are paired up in a meaningful way- there are infinitely many mappings $G$". On the other hand, a design objective as a dictionary \cite{b33} learning is not practically possible due to the complexity of the parameter space and scale. Because manipulating dictionary metrics in training and inference stages is computationally costly \cite{b21}. 

The idea is to build an embedding vector training process without explicitly using an embedding vector. As well as the model architecture must support proper discriminative loss function to be trained in the adversarial phase.

The concept of metrics learning in deep learning \cite{b23} for similarity search is the optimal way to train Because, in theory, such a model uses sparse categorical cross-entropy as the objective function.  As stated in the objective section, this objective makes encoding vector distribution estimates for each class so that their negative log-likelihood is as minimal as possible. In essence, this makes encoding vectors of the same class drop under the same distribution and achieve distribution clustering by preserving proximity. 

As stated in Deep-TEN \cite{b11} CNN models with residual connection from low level feature layers to high level feature layers are suitable for tasks like material and texture extraction. In \cite{b11} they have empirically proven this point by using the ResNet model to classify garments based on texture. However, the DensNet \cite{b18} model is a perfect candidate for this because it concatenates all features in previous modules, in the current module, and the transition layer is weighted between different layers to generate more complex features than features generated by ResNet \cite{b19} on a particular resolution. This makes the model have a much stronger gradient flow. Also, in \cite{b18} the width of the layer being proportional to the growth rate makes the feature extractor of the style encoder have to have a much smaller parameter space.

\paragraph{\textbf{Generator} -}
Generator models are mainly a combination of content encoder, style encoder and decoder. Here the decoder model is a sequential set of up-sampling CNN modules like the U-Net with skip connection to content encoder sub-model. The network architecture is inspired by the StyleGAN \cite{b7} synthesis network. Instead of using a single mapping network \cite{b7} to generate features from the inputs here we implement two separate encoder models as shown in Fig.~\ref{fig:app_2_gan}.

The decoder sub-model in the generator starts from the latent vector space generated only by the style encoder. The generator model is fed by the low-level feature from the content encoder using skip connection. The targethere is to make decoder sensitivity style encoded high-level features while using low-level features to generate content on the image. Therefore, the final image will not be affected by the exact shapes and colors of the content image and what objects are present in the style image.

\begin{figure}[!t]
 \centering
  \includegraphics[width=\linewidth]{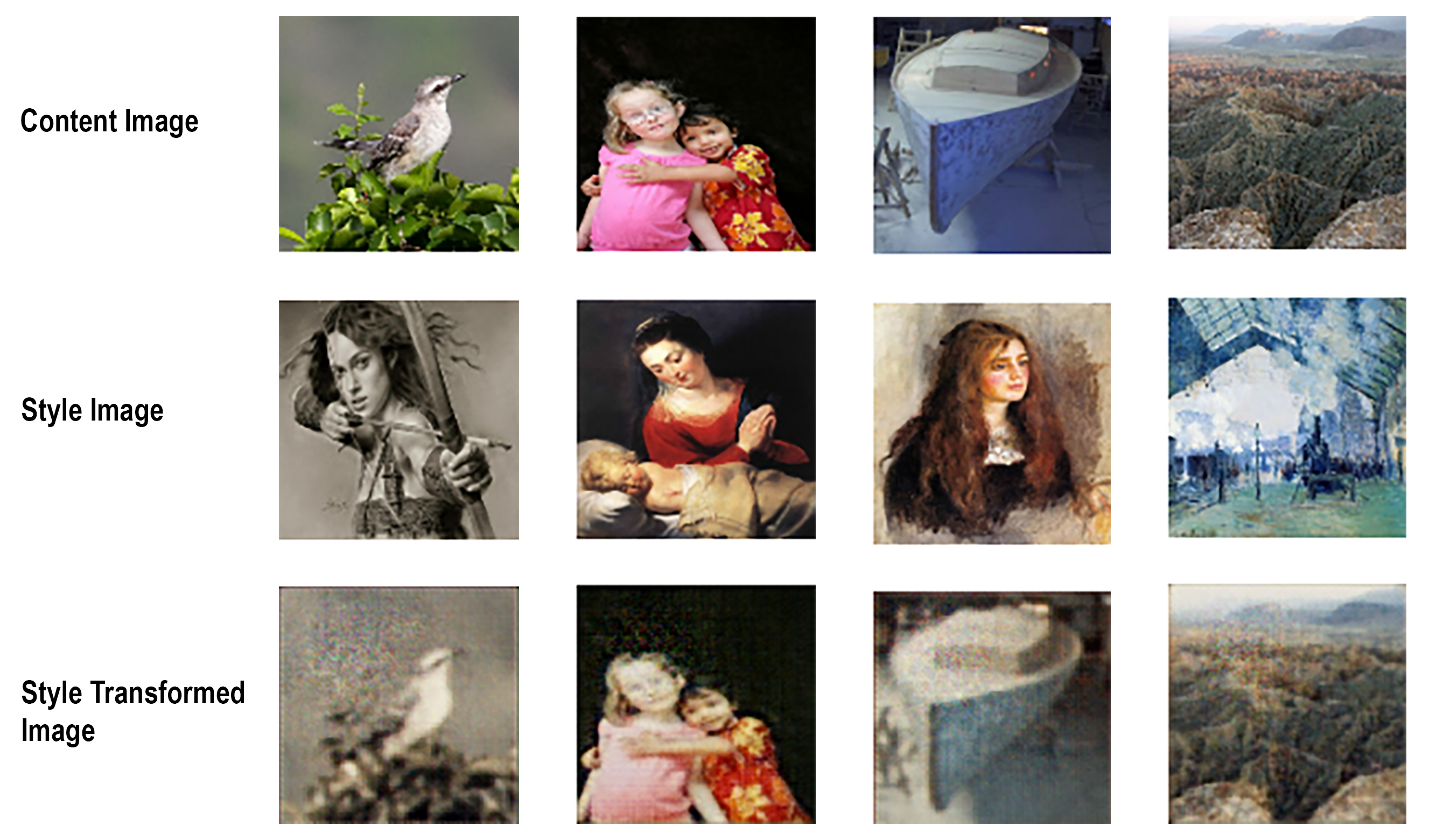}
  \caption{Approach \#2 GAN model evaluation image matrix.}
  \label{fig:app_2_eval}
\end{figure}

\section{Model Evaluation}
\subsection{Results of approach \#1}
Despite having two discriminative models the discriminator in total and individually tended to converge into an equilibrium point in the training as we experimented. Furthermore, the generator shows less variety than the CNN-based NST model in loss, especially l2 loss in experiment results.

The resulting generated images tend to have a much higher resemblance to the respective art style compared to the CNN-based NST model, as illustrated, especially in the third and fourth rows of Fig.~\ref{fig:app_1_eval}. Especially in generated images, the presence of content color palette dropping is insignificant. Furthermore, the introduction of new artifacts to generated images are not prominent with the new model.

\subsection{Results of approach \#2}

For the GAN training process we used two separate datasets for style images and content images as stated in the method section. The complete adversarial training process on style discriminator, content discriminator and the generator converge to an optimal point by minimizing loss function related to each instead of the discriminative loss oscillating over equilibrium level. Also, generative loss and content encoder loss happened to have smooth loss graphs relative to the style encoder. Here in a batch the presence of class is not uniformly distributed.

Even Though we had adopted the matrix learning to calculate the model loss, the sample set does not always contain samples from each class in the same order. Under this method the model overfitting and tendency to model collapsing reduce significantly than the orders static sampling method.

The generated images have not changed radically like in \cite{b28} model inference, but the generated images are imposed with texture and some color effect from the style image as illustrated in Fig.~\ref{fig:app_2_eval}. As proposed, the model is supposed to generate an image which will appear as done by the same artist rather than seeming to have all features from a given style image. Considering the above mentioned point, the model we introduce does a quite good job in understanding style image texture and imposing that on the content image without introducing extrinsic features from style image as artifacts. The slight amount of noise and blur represented in generated images is due to the downscaling of input images. 

\section{Discussion}
\subsection{Lessons Learned}
In the first approach, we have mainly targeted building a GAN for NST that supports more than one style. The generating image shows significantly less content color palette changes than the CNN base NST models because of the adversarial training. Texture discontinuity and blurriness of the generated image were significantly affected by the severe downsampling of the images’ pixel space because the downsampling operation could drop interpixel correlations present in the pixel space.

From the second approach results, we can specify that the style extraction by using metric learning emphasizes general features for a style class. Consequently, the generated images have texture and similar semantic appearance features from the style image.
On the other hand, the generated image will not have striking features such as rapid contract change and sharp edge as in the reference style image because of style normalization, in a broader sense. 

In this paper we introduce a comprehensive architecture for the GAN which has two parallel discriminator headers and generators that takes 6 channel tensor inputs. Because this model has a larger parameter space than the general GAN model, the training process is slower than average, but the inference time has not changed significantly. On the other hand, the model proposed in the second approach has even less parameter space than the general GAN model. However, training the encoder model parameter set sharing between generator and discriminator caused an increase in the expected return time. Although, the inference is much faster than approach \#1 due to the smaller model size and fewer FLOPS.

\subsection{Open Challenges and Future Research Directions}
The model proposed in approach \#2 is not capable of clearly introducing striking features such as outline markings and rapid color gradient patches into the generated image even though they appear in the style image. Hence, the proposed model is only fine-tuned on very low-resolution images (128, 128, 3), where the severe downscaling of input images caused edges of the images to blur and be discontinued. Therefore, input images being in very low resolution could cause the issue of not introducing striking features.

The target sensor replacement method introduced in \cite{b20} paper is trained in the Eigenspace of the related pixel space. It ensures that the decoder is fed with essential details about style in the bottleneck layer. Also, we can have more control over the latent space, implying that we can build a model that controls the style transfer \cite{b30, b35} under rGAN architecture. Building Multimodal architecture that serves like the model introduced in this paper and with aleatoric uncertainty of generating image would be like variational autoencoder \cite{b22} among autoencoders \cite{b13}.

\section{Conclusion}
In this paper, we introduced two novel semi-supervised learning approaches of neural artistic style transformation. The first approach used a GAN-based style transfer model capable of transferring images into more than one style. In the second approach, we introduced  neural style transfer that understands how an actual artist would paint a given image in his style. The adversarial network and adversarial training process introduced in the second approach would be applicable to many ambiguous tasks and train on a lower parameter space than the conventional GAN model.

\section*{Acknowledgment}

We thank L.T.N. Wickremasinghe, T.T. Jayasekera and P.G. Amanda for the support on this project. 


\bibliographystyle{model1-displa-num-names}
\bibliography{refs}

\end{document}